\documentclass[10pt,twocolumn,letterpaper]{article}

\usepackage{cvpr}              

\usepackage[dvipsnames]{xcolor}
\usepackage{multirow}

\definecolor{cvprblue}{rgb}{0.21,0.49,0.74}
\usepackage[pagebackref,breaklinks,colorlinks,citecolor=cvprblue]{hyperref}
\usepackage[accsupp]{axessibility} 

\def\paperID{13619} 
\def\confName{CVPR}
\def\confYear{2024}

\title{MVIP-NeRF: Multi-view 3D Inpainting on NeRF Scenes via Diffusion Prior}

\author{Honghua Chen \qquad Chen Change Loy \qquad  Xingang Pan \\
S-Lab, Nanyang Technological University\\
{\tt\small honghua.chen@ntu.edu.sg \qquad ccloy@ntu.edu.sg \qquad xingang.pan@ntu.edu.sg}
}

\begin{document}

\twocolumn[{%
\renewcommand\twocolumn[1][]{#1}%
\maketitle
\begin{center}
    \centering
    \includegraphics[width=\linewidth]{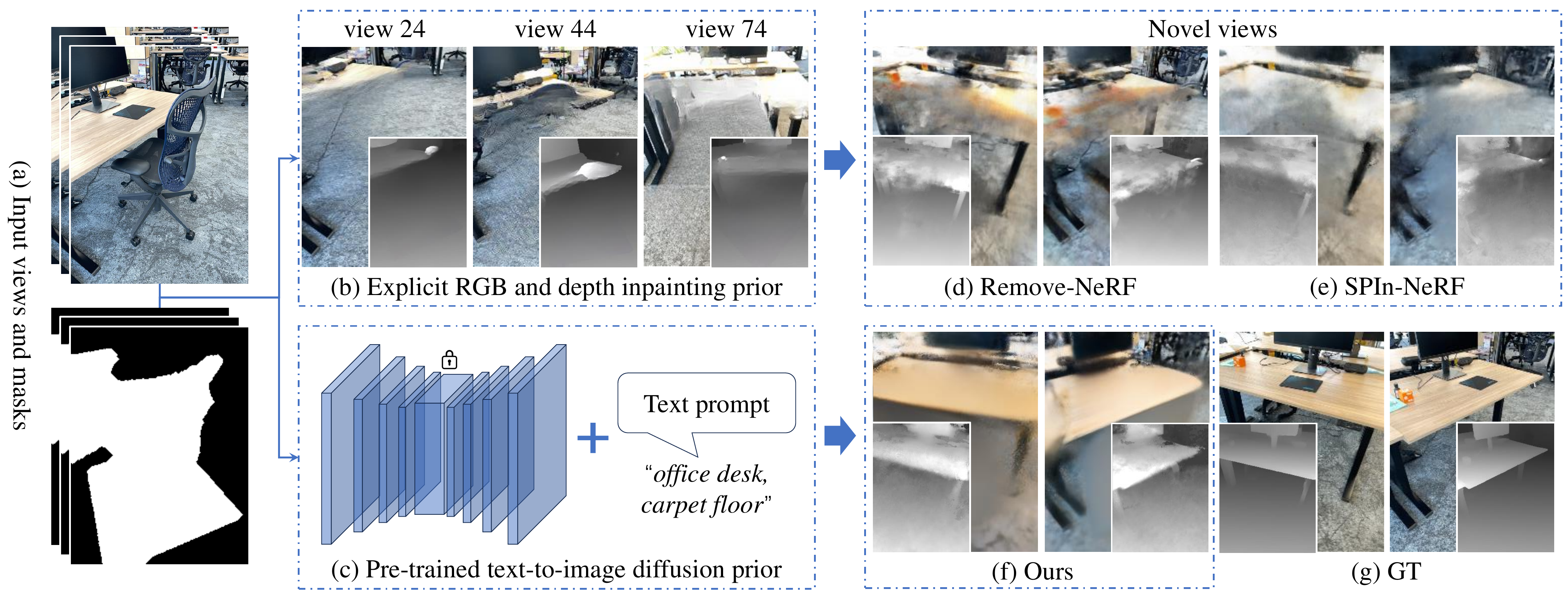}
    \captionof{figure}{Comparison of our MVIP-NeRF with two state-of-the-art approaches, Remove-NeRF~\cite{weder2023removing} and SPIn-NeRF~\cite{mirzaei2023spin}. Existing methods heavily depend on explicit RGB and depth inpainting results. This type of inpainting prior frequently shows inconsistency, inaccuracy, and misalignment to a certain degree (sub-figure (b)). In contrast, our approach implicitly exploits the diffusion prior (sub-figure (c)), resulting in more faithful and consistent results, in terms of both appearance and geometry.}\label{fig:teaser}
\end{center}%
}]

\maketitle
\begin{abstract}
Despite the emergence of successful NeRF inpainting methods built upon explicit RGB and depth 2D inpainting supervisions, these methods are inherently constrained by the capabilities of their underlying 2D inpainters. This is due to two key reasons: (i) independently inpainting constituent images results in view-inconsistent imagery, and (ii) 2D inpainters struggle to ensure high-quality geometry completion and alignment with inpainted RGB images.

To overcome these limitations, we propose a novel approach called MVIP-NeRF that harnesses the potential of diffusion priors for NeRF inpainting, addressing both appearance and geometry aspects.
MVIP-NeRF performs joint inpainting across multiple views to reach a consistent solution, which is achieved via an iterative optimization process based on Score Distillation Sampling (SDS).
Apart from recovering the rendered RGB images, we also extract normal maps as a geometric representation and define a normal SDS loss that motivates accurate geometry inpainting and alignment with the appearance.
Additionally, we formulate a multi-view SDS score function to distill generative priors simultaneously from different view images, ensuring consistent visual completion when dealing with large view variations.
Our experimental results show better appearance and geometry recovery than previous NeRF inpainting methods.
\end{abstract}

\section{Introduction}
Neural Radiance Fields (NeRFs)~\cite{MildenhallSTBRN20} inpainting involves the removal of undesired regions from a 3D scene, with the objective of completing these regions in a contextually coherent, visually plausible, geometrically accurate, and consistent manner across multiple views. 
This form of 3D editing holds significant value for diverse applications, including 3D content creation and virtual/augmented reality.

Inpainting on NeRF scenes presents two intricate challenges: 
(i) how to ensure that the same region observed in multiple views is completed in a consistent way, especially when the view changes significantly; 
and (ii) inpainting must address not only the 2D appearance of NeRFs but also yield geometrically valid completion.

Several NeRF inpainting techniques have been developed to address specific aspects of these challenges~\cite{mirzaei2023spin,weder2023removing,mirzaei2023reference,zhou2023repaint,yin2023or,liu2022nerf}. The majority of these approaches heavily rely on explicit RGB and depth inpainting priors, often employing 2D inpainters like LaMa~\cite{suvorov2022resolution} to independently inpaint all views and subsequently address the multi-view inconsistency.
For example, SPIn-NeRF~\cite{mirzaei2023spin} and InpaintNeRF360~\cite{wang2023inpaintnerf360} incorporate a perceptual loss within masked regions to account for low-level inconsistency, but the perceptual-level inconsistency still cannot be fully addressed (see from Fig.~\ref{fig:teaser} (b)(e)).
Another approach involves preventing inconsistent and incorrect views from being used in NeRF optimization. To achieve this, Weder et al.~\cite{weder2023removing} introduce uncertainty variables to model the confidence of 2D inpainting results, facilitating automated view selection.
As a simpler alternative, Mirzaei et al.~\cite{mirzaei2023reference} propose to use a single inpainted reference view to guide the entire scene inpainting process. However, this method is difficult to adapt to scenes with large view variations and requires non-trivial depth alignment. In summary, these methods remain constrained by the capabilities of underlying 2D inpainters. Besides, they share the common limitation of neglecting the correlation between
inpainted RGB images and inpainted depth maps, resulting in less pleasing geometry completion.

In this work, we are interested in addressing these challenges via a new paradigm.
Instead of employing 2D inpainting independently for each view, we believe that ideally, the inpainting at different views should work jointly to reach a solution that \textit{i)} fulfills the 2D inpainting goal at each view and \textit{ii)} ensures 3D consistency.
Fortunately, 2D diffusion models present an ideal prior for achieving this goal.
While recent advances like DreamFusion~\cite{poole2022dreamfusion} have demonstrated their capability in 3D generation, the adaptation of diffusion priors to tackle the NeRF inpainting problem remains an untapped area.

To this end, we present \textit{MVIP-NeRF}, a novel approach that performs multiview-consistent inpainting in NeRF scenes via diffusion priors.
Given an RGB sequence and per-frame masks specifying the region to be removed, we train a NeRF using a reconstruction loss in the observed region and an inpainting loss in the masked region.
The inpainting loss is based on the Score Distillation Sampling (SDS)~\cite{poole2022dreamfusion} that attempts to align each rendered view with the text-conditioned diffusion prior.
This approach allows our model to progressively fill the missing regions in the shared 3D space, thus the inpainting goal at multiple views can work jointly to reach a consistent 3D inpainting solution. 
To further ensure a valid and coherent geometry in the inpainted region, we also adopt diffusion priors to optimize the rendered normal maps.
In addition, observing that the stochasticity of SDS often leads to a sub-optimal solution under large view variations, we formulate a multi-view score distillation, which ensures that each score distillation step takes into account multiple views that share the same SDS parameters.
This achieves improved consistency and sharpness within the filled regions when the view changes significantly. We summarize our contributions as follows:

(i) A diffusion prior guided approach for high-quality NeRF inpainting, achieved without the need for explicit supervision of inpainted RGB images and depth maps.

(ii) An RGB and normal map co-filling scheme with iterative SDS losses that can simultaneously complete and align the appearance and geometry of NeRF scenes.

(iii) A multi-view score function to enhance collaborative knowledge distillation from diffusion models, avoiding detail blurring when dealing with large view variations.

(iv) Extensive experiments to show the effectiveness of our method over existing NeRF inpainting techniques.

\section{Related Work}
\subsection{NeRF Inpainting}
The use of NeRFs~\cite{MildenhallSTBRN20} for representing 3D scenes has enabled high-quality, photorealistic novel view synthesis. Despite this, only a limited number of studies have delved into the task of object removal or inpainting from pre-trained NeRF models.
Early approaches such as EditNeRF~\cite{liu2021editing}, Clip-NeRF~ \cite{wang2022clip} and LaTeRF~ \cite{mirzaei2022laterf}, introduced methods to modify objects represented by NeRFs. However, the efficacy of these approaches is largely limited to simple objects rather than scenes featuring significant clutter and texture. Object-NeRF~ \cite{yang2021learning} supports the manipulation of multiple objects, like moving, rotating, and duplicating, but does not carefully handle the inpainting scenario. More closely, Instruct-NeRF2NeRF~\cite{haque2023instruct} proposes to use an image-conditioned diffusion model to facilitate text-instructed NeRF scene stylization.

NeRF-In~\cite{liu2022nerf}, Remove-NeRF~\cite{weder2023removing}, SPIn-NeRF~\cite{mirzaei2023spin}, and InpaintNeRF360~\cite{wang2023inpaintnerf360} are most closely related to our method. All of these approaches use RGB and depth priors from 2D image inpainters to inpaint NeRF scenes. The main difference among them is how they resolve view inconsistencies.
Remove-NeRF~\cite{weder2023removing} tackles inconsistencies by adaptively selecting views based on the confidence of the 2D inpainting results. Following it, a more straightforward scheme is to only use a single inpainted reference image to guide NeRF scene inpainting~\cite{mirzaei2023reference}. However, this method necessitates tedious and exact depth alignment. SPIn-NeRF~\cite{mirzaei2023spin} and InpaintNeRF360~\cite{wang2023inpaintnerf360} employ a perceptual loss within inpainted regions to account for the inconsistencies between different views. 
In contrast to these methods, we propose to inpaint NeRF scenes through an iterative optimization process that distills appearance and geometry knowledge from the pre-trained diffusion model. Consequently, our approach attains a more consistent and realistic representation of the entire NeRF scene, without requiring explicit RGB or depth supervision.
It is worth noting that a recent work, RePaint-NeRF~\cite{zhou2023repaint}, also leverages a pre-trained diffusion model for NeRF painting. However, its primary focus is on object replacement within a NeRF scene. This task differs from ours in that it does not necessitate considering the coherence with the local context or ensuring high-quality geometry filling. 

\subsection{Diffusion Priors}
Recently, we have witnessed remarkable advancements in the field of image generation, driven by the evolution of diffusion models~\cite{Diff15, DDPM, ScoreBased, DiffBeatGAN}.
These models excel by progressively removing noise from Gaussian distributions using a UNet noise predictor, enabling the generation of high-quality images that align well with the training data.
By training on large-scale text-image pairs~\cite{Schuhmann2022LAION5BAO}, diffusion models have gained unprecedented success in text-to-image generation, with Stable Diffusion~\cite{rombach2022high} as a phenomenal example.
Therefore, many efforts have been made to explore the use of diffusion models as priors for a range of image restoration tasks such as super-resolution, colorization, inpainting, and deburring, \etc~\cite{kawar2022denoising, wang2023ddnm, lin2023diffbir, wang2023exploiting, li2023diffusion}.

Beyond their use in 2D tasks, diffusion priors have also seen successful applications in 3D generation.
A pioneering work in this direction is Dreamfusion~\cite{poole2022dreamfusion}, which leverages multiview 2D diffusion priors for 3D generation via an SDS loss,  a concept derived from the distillation process of Imagen~\cite{saharia2022photorealistic}.
This approach gets rid of the need for large amount of 3D training data and thus has been widely adopted in subsequent text-to-3D synthesis endeavors such as MakeIt3D~\cite{tang2023make}, Magic3D~\cite{lin2023magic3d}, Fantasia3D~\cite{chen2023fantasia3d}, and ProlificDreamer~\cite{wang2023prolificdreamer}.
The SDS loss can not only synthesize objects but also edit existing ones, as studied in Latent-NeRF~\cite{metzer2023latent}, Vox-E~\cite{sella2023vox}, and AvatarStudio~\cite{mendiratta2023avatarstudio}.
Unlike these works that aim to create or edit 3D objects, our work is targeted to inpaint undesired regions to be coherent with the context for NeRF scenes.

\begin{figure*}[t]
    \centering
    \includegraphics[width=\linewidth]{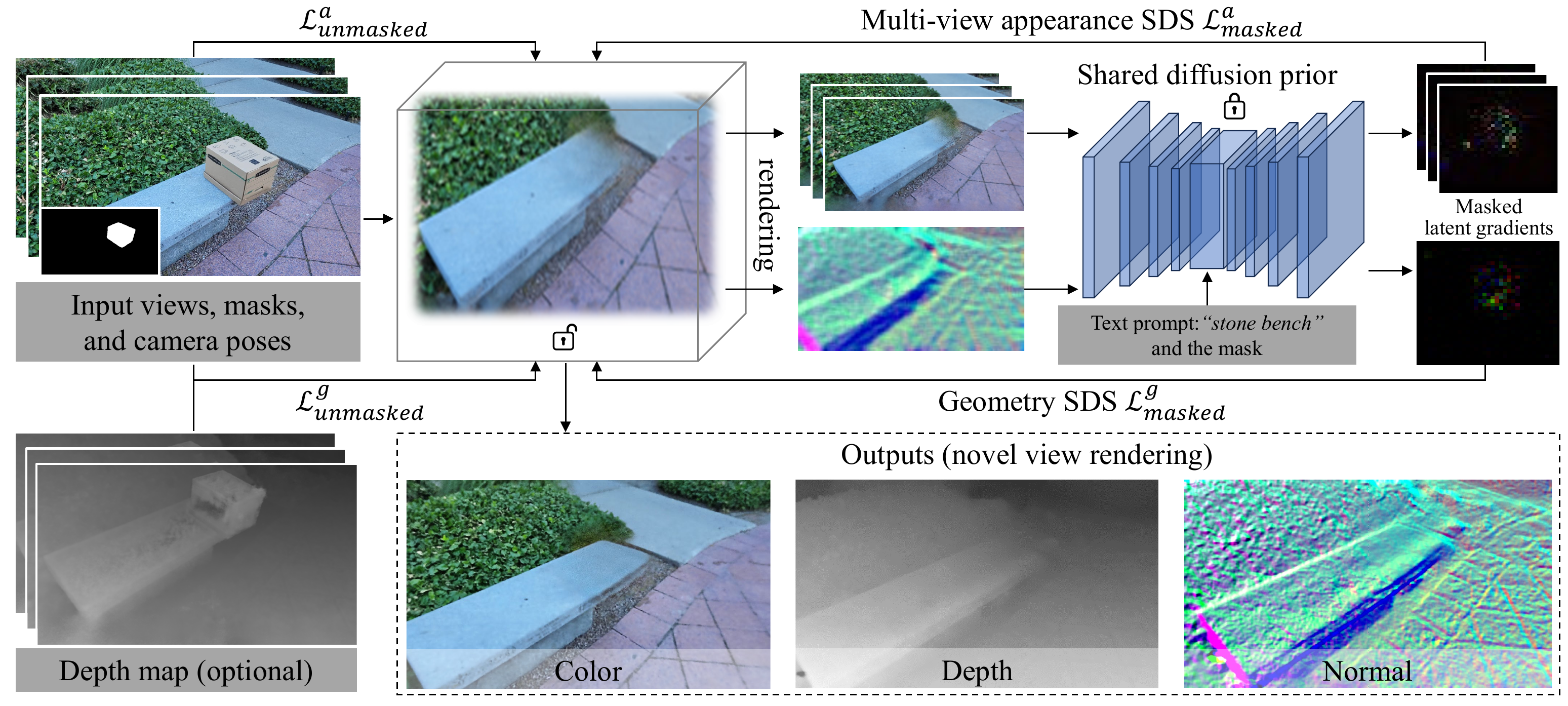}
    \caption{Method overview. Given posed RGB images with corresponding masks, depth maps (optional), and a text description, MVIP-NeRF can faithfully recover plausible textures and accurate surface detail. In the optimization process, for unmasked regions, we employ direct pixel-wise RGB and depth reconstruction losses. For masked regions, we introduce an RGB and normal map co-filling approach, utilizing SDS losses. This approach iteratively completes and aligns the appearance and geometry of NeRF scenes without the need for explicit supervision. Furthermore, we implement a multi-view scoring mechanism within the diffusion process to effectively handle significant variations in viewpoints. Finally, novel views can be rendered from the NeRF scene, where the object has been removed.}
    \label{fig:pipeline}
\end{figure*}

\section{Method}
In this section, we provide a brief introduction to NeRF and SDS, followed by the formulation of our problem setting.

\subsection{Preliminary}
\noindent\textbf{Neural Radiance Fields.} NeRFs~\cite{MildenhallSTBRN20} encodes a 3D scene, by a function $g$ that maps a 3D coordinate $\mathbf{p}$ and a viewing direction $\mathbf{d}$ into a color value $\mathbf{c}$ and a density value $\sigma$. The function $g$ is a neural network parameterized by $\theta$, so that $g_{\theta}:(\gamma(\mathbf{p}),\gamma(\mathbf{d}))\mapsto(\mathbf{c}, \sigma)$, where $\gamma$ is a positional encoding. Each expected pixel color $\hat{C}(\mathbf{r})$ is rendered by casting a ray $\mathbf{r}$ with near and far bounds $t_n$ and $t_f$. Typically, we divide $[t_n, t_f]$ into $N$ sections ($t_1, t_2, ..., t_N$) along a ray $\mathbf{r}$ and then compute the pixel color by $\hat{C}(\mathbf{r})=\sum_{i=1}^{N}\mathbf{c}_i^{*}$. The  weighted color $\mathbf{c}_i^{*}$ of a 3D point is computed by $\mathbf{c}_i^{*}=w_i\mathbf{c}_i$, where $\quad w_i=T_i(1-\mathrm{exp}(-\sigma_i\delta_i))$,  $T_i=\mathrm{exp}(-\sum_{j=1}^{i-1}\sigma_j\delta_j)$, and $\delta_i=t_{i}-t_{i-1}$. Therefore, the NeRF reconstruction loss can be formulated as 
\begin{equation}
\label{eq:reco_a}
\mathcal{L}^{a}=\sum_{\mathbf{r}\in R}||\hat{C}(\mathbf{r})-C(\mathbf{r})||^{2},
\end{equation}
where $\hat{C}(\mathbf{r})$ represents the rendered color blended from $N$ samples, $R$ is a batch of rays randomly sampled from the training views, and $C(\mathbf{r})$ corresponds to the ground-truth pixel color. If equipped with the ground-truth depth information, we can add another reconstruction loss to further optimize the geometry of NeRF scenes~\cite{deng2022depth}:
\begin{equation}
\label{eq:reco_g}
\mathcal{L}^{g} = \sum_{\mathbf{r}\in R}||\hat{D}(\mathbf{r})-D(\mathbf{r})||^{2},
\end{equation}
where $\hat{D}(\mathbf{r})$ represents the rendered depth/disparity and $D(\mathbf{r})$ corresponds to the ground-truth pixel depth.

\noindent\textbf{Score distillation sampling.}
SDS~\cite{poole2022dreamfusion} enables the optimization of any differentiable image generator, \eg, NeRFs or the image space itself.
Formally, let $\mathbf{x}=g(\theta)$ represent an image rendered by a differentiable generator $g$ with parameter $\theta$, then SDS minimizes density distillation loss~\cite{oord2018parallel} which is essentially the KL divergence between the posterior of $\mathbf{x} = g(\theta)$ and the text-conditional density $p_\phi^{\omega}$:
\begin{equation} 
\begin{split}
\mathcal{L}_{\tt{Dist}}(\theta) =\mathbb{E}_{t,\boldsymbol{\epsilon}}\big[w(t)\,D_{\tt{KL}}\big( q\big(\mathbf{x}_t|\mathbf{x}\big) \,\|\, p_\phi^{\omega}(\mathbf{x}_t; y,t)\big) \big],
    \label{eq:KL}
\end{split}
\end{equation}
where $w(t)$ is a weighting function, $y$ is the text embedding, and $t$ is the noise level.
For an efficient implementation, SDS updates the parameter $\theta$ by randomly choosing timesteps $t\sim \mathcal{U}(t_{\tt{min}}, t_{\tt{max}})$ and forward $\mathbf{x}=g(\theta)$ with noise $\boldsymbol\epsilon\sim \mathcal{N}(\mathbf{0},\mathbf{I})$ to compute the gradient as: 
\begin{align}\label{eq:sds}
    \nabla_\theta \mathcal{L}_{\tt{SDS}}(\theta) = \mathbb{E}_{t,\boldsymbol{\epsilon}}\left[ w(t) \big(\boldsymbol{\epsilon}_\phi^{\omega}(\mathbf{x}_t; y,t) - \boldsymbol{\epsilon}\big)\frac{\partial \mathbf{x}}{\partial \theta} \right].
\end{align}

\subsection{Problem formulation and overview}
Given a set of RGB images, $\mathcal{I} = \{ I_i \}_{i= 1}^{n}$, with corresponding 3D poses $\mathcal{G} = \{ G_i \}_{i= 1}^{n}$, 2D masks $\mathcal{M} = \{ m_i \}_{i= 1}^{n}$, and a text description $y$, our goal is to produce a NeRF model for the scene. This NeRF model should have the capability to generate an \textit{inpainted} image from any novel viewpoint. 
In general, we address unmasked and masked regions separately, following this general formulation:
\begin{equation}
\label{eq:formulation}
\mathcal{L}=\mathcal{L}_{\mathrm{unmasked}}^{a} + \lambda_1\mathcal{L}_{\mathrm{unmasked}}^{g} + \lambda_2\mathcal{L}_{\mathrm{masked}}^{a} + \lambda_3\mathcal{L}_{\mathrm{masked}}^{g}.
\end{equation}
The entire process is visualized in Figure~\ref{fig:pipeline}. Specifically, for unmasked regions, we utilize pixel-wise color (Eq.~\ref{eq:reco_a}) and depth (Eq.~\ref{eq:reco_g}) reconstruction loss. For masked regions, we first render an RGB image and a normal map from the NeRF scene. Then, a latent diffusion model is employed as the appearance and geometry prior. Rather than directly utilizing inconsistent 2D inpainting results as supervisions and resolving these inconsistencies \textit{post hoc}, we employ two SDS losses to compute a gradient direction iteratively for detailed and high-quality appearance and geometry completion. To further enhance consistency for large-view motion, we introduce a multi-view score function. This function ensures that multiple sampled views share the knowledge distilled from the diffusion models, thereby promoting cross-view consistency. Next, we will explain how to define $\mathcal{L}_{\mathrm{masked}}^{a}$ and $\mathcal{L}_{\mathrm{masked}}^{g}$, and how to extend these concepts to a multi-view version.

\subsection{Appearance Diffusion Prior}
We have noticed that independently inpainting individual images does not guarantee a consistent completion of the same region observed from multiple views. Sometimes, the inpainted results may even be incorrect. Therefore, instead of relying on explicit inpainting images, we incorporate a diffusion prior. We treat the inpainting task as a progressive denoising problem, which not only ensures view consistency but also enhances the visual realism of the completed scenes.
To be more specific, we define the appearance SDS within the latent space of Stable Diffusion:
\begin{equation}\label{eq:appearance_sds}
    \nabla_\theta \mathcal{L}_{\mathrm{masked}}^{a} = w(t) \big(\boldsymbol{\epsilon}_\phi^{\omega}(\mathbf{z}_t; m,y,t) - \boldsymbol{\epsilon}\big)\frac{
    \partial \mathbf{z}}{\partial \mathbf{x}}\frac{
    \partial \mathbf{x}}{\partial \theta},
\end{equation}
where the noisy latent $\mathbf{z}_t$ is obtained from a novel view rendering $\mathbf{x}$ by Stable Diffusion encoder and $m$ is the corresponding mask. It is important to note that we only backpropagate the gradient for the masked pixels.
The range of timesteps $t_{\tt{min}}$ and $t_{\tt{max}}$ are chosen to sample from not too small or large noise levels and the text prompt $y$ should align with the missing regions.
In this work, we use the stable-diffusion-inpainting model \cite{rombach2022high} as our guidance model.

\subsection{Geometry Diffusion Prior}
In NeRF scenes, aside from appearance, achieving accurate geometry is a crucial component. Previous NeRF inpainting methods employ inpainted depth maps as an additional form of guidance for the NeRF model. However, we have observed that these inpainted depth maps often lead to visually unsatisfactory results and exhibit poor alignment with RGB images (see from Fig.~\ref{fig:teaser} (b)). Consequently, this approach tends to be less effective in achieving high-quality geometry restoration in the masked areas.

In our work, we have two observations: (i) text-to-image diffusion models have a strong shape prior due to their training on diverse objects, and (ii) surface normals clearly reveal the geometric structures.
Both observations encourage us to decouple the generation of geometry from appearance while further exploiting the potential geometry information from the diffusion prior. More specifically, considering the current NeRF function as $g(\theta)$, we generate a normal map $\mathbf{n}$ by rendering it from a randomly sampled camera pose.
To update $\theta$, we again employ the SDS loss that computes the gradient w.r.t. $\theta$ as:
\begin{align}
    \nabla_\theta \mathcal{L}_{\mathrm{masked}}^{g} = w(t) \big(\boldsymbol{\epsilon}_\phi^{\omega}(\mathbf{z}_t; m,y,t) - \boldsymbol{\epsilon}\big)\frac{
    \partial \mathbf{z}}{\partial \mathbf{n}}\frac{
    \partial \mathbf{n}}{\partial \theta}.
\end{align}

\begin{figure}[t]
    \centering
    \includegraphics[width=\linewidth]{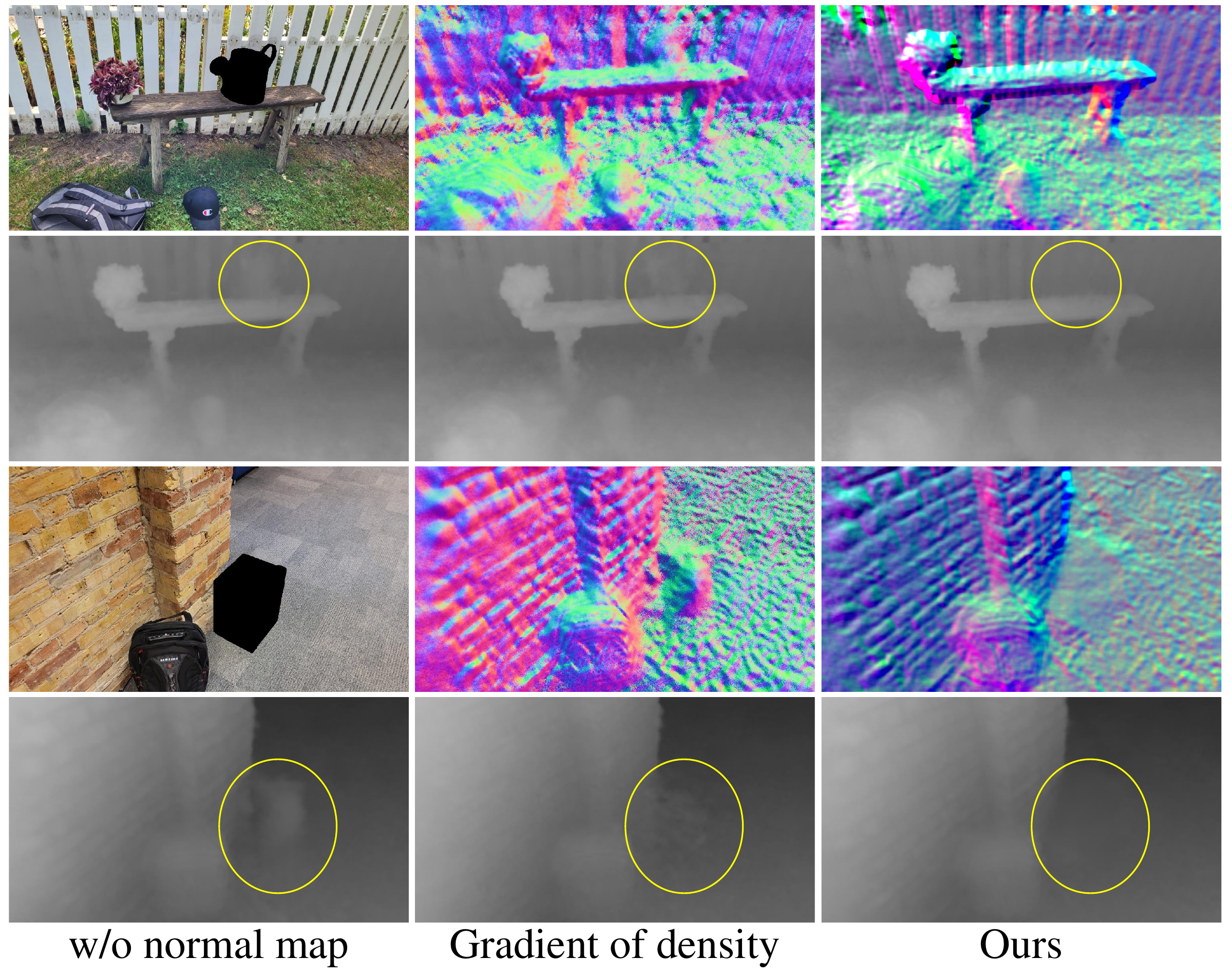}
    \caption{Effect of different normal map generation methods. In the first column, we present the input image with a mask (black region) and the depth map generated by NeRF, optimized with unmasked pixels. The second column displays the normal map derived from the density field gradient and the corresponding optimized depth map. The final column highlights the improved accuracy and reliability of geometry reconstruction achieved through the use of a smoothed normal field.}
    \label{fig:normal}
\end{figure}

\noindent\textbf{Smoothed normal map generation.}
Surface normals are commonly derived by computing the gradient of the density field $\sigma$ with respect to sampled positions. Nevertheless, the computed normals may exhibit some degree of noise, leading to an unclear geometric context. This, in turn, results in instability when generating geometry within the masked region, as demonstrated in Figure~\ref{fig:normal}. Considering the readily available camera parameters and depth map, we introduce to calculate the smoothed surface normal from the depth map, treating it as a differentiable plane fitting problem.

Specifically, we denote $(u_{i},v_{i})$ as the location of pixel $i$ in the 2D image. Its corresponding location in 3D space is $(x_{i},y_{i},z_{i})$. We adopt the camera intrinsic matrix to compute $(x_{i},y_{i},z_{i})$ from its 2D coordinates $(u_{i}, v_{i})$, where $z_{i}$ is the depth and given.
Based on the assumption that pixels within a local neighborhood of pixel $i$ lie on the same tangent plane, we then build the tangent plane to compute the surface normal of pixel $i$. In particular, we search the $K$ ($K=9$ in default) nearest neighbors in 3D space and calculate the surface normal estimate $\mathbf{n}$ based on these neighboring pixels on the tangent plane. The surface normal estimate $\mathbf{n}$ should satisfy the linear system of equations:
\begin{equation}
\mathbf{A}\mathbf{n} = \mathbf{b},~~\rm{subject~to}~~\Vert\mathbf{n}\Vert^{2}_{2} = 1,
\label{eq:An}
\end{equation}
where $\mathbf{A}\in R^{K\times3}$ is a matrix grouped by neighboring pixels and $\mathbf{b} \in R^{K\times1}$ is a constant vector. Finally, we obtain the surface normal by minimizing $\Vert\mathbf{An}-\mathbf{b}\Vert^{2}$ whose least square solution has the closed form.

\begin{figure}[t]
    \centering
    \includegraphics[width=\linewidth]{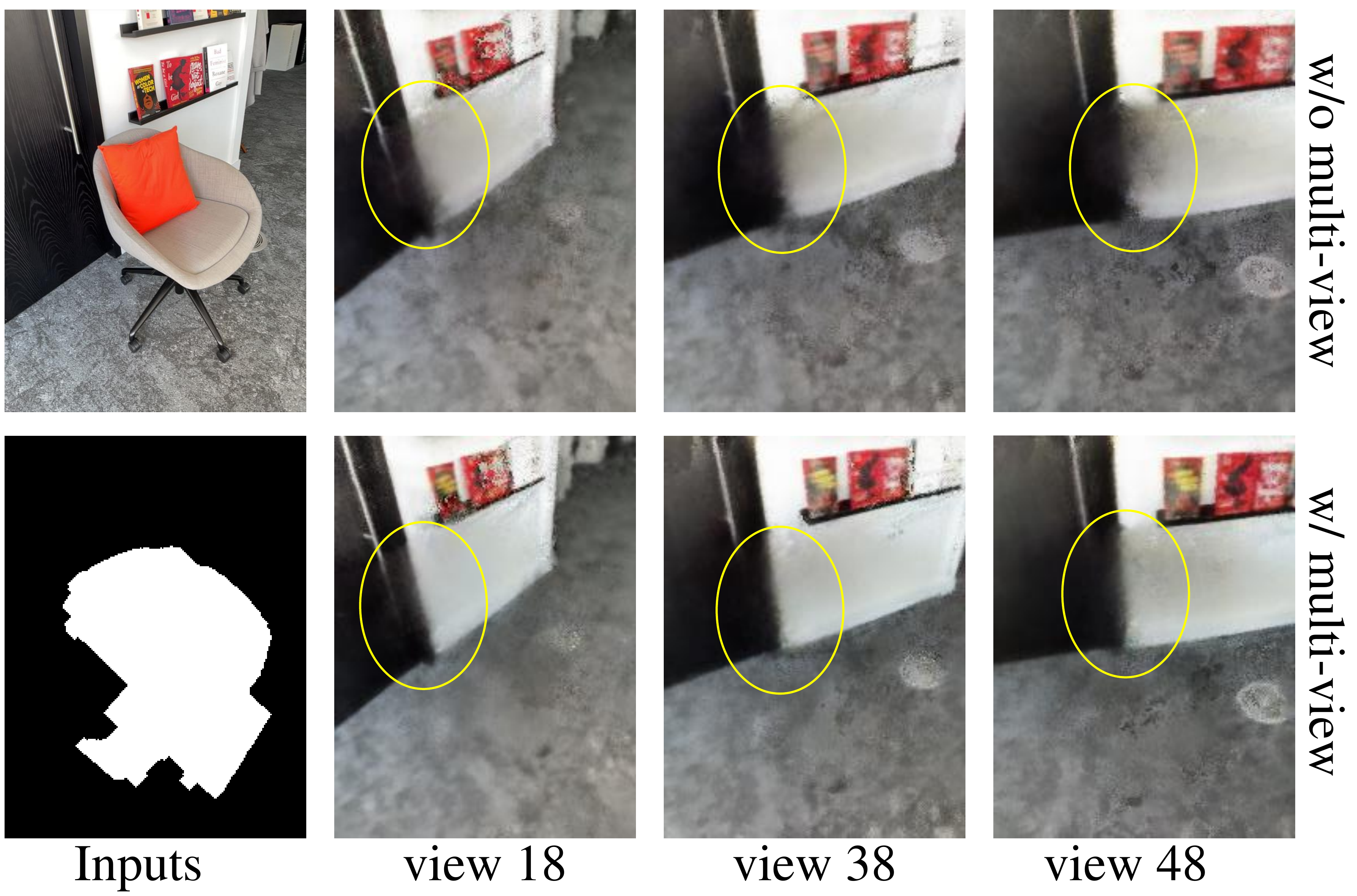}
    \caption{Effect of multi-view score distillation. The first row shows inpainting results without the multi-view score, while the second row shows the results with the multi-view score ($N=5$).}
    \label{fig:large-view}
\end{figure}

\subsection{Multi-view Score Distillation }
While our method consistently produces reliable results, it is worth noting that some blurring may occur in scenarios with large view variation, as shown in Fig.~\ref{fig:large-view}. The primary reason behind this may be that previous gradient updating, which relies on single-view information, does not adequately account for cross-view information.
To this end, we define a multi-view distillation score function, which enhances the correlation in the recovery of each view.
Given $N$ viewpoints, we accordingly render $N$ images, denoted as $\{\mathbf{x}^1,...,\mathbf{x}^N\}$. 
Naturally, we can compute a multi-view score as:
\begin{equation}
\begin{split}
    \nabla_\theta \mathcal{L}_{\mathrm{masked}}^{ma}= 
    \sum_{i=1}^{N}\big(w(t) (\boldsymbol{\epsilon}_\phi^{\omega}(\mathbf{z}_t^i; m^i,y,t) - \boldsymbol{\epsilon}^i)
    \frac{\partial \mathbf{z}^i}{\partial \mathbf{x}^i}\frac{
    \partial \mathbf{x}^i}{\partial \theta}\big).
\label{eq:multi_appearance_sds}
\end{split}
\end{equation}
Note that the noise estimator and the noise level $t$ are shared for all $N$ images. Intuitively, this function implies that when updating $\theta$, we take into account the interactions with other sampled views, thereby promoting view consistency.

\noindent \textbf{Final loss function.}
Finally, we replace the $\mathcal{L}_{\mathrm{masked}}^{a}$ with its multi-view version, and jointly train the loss as:
\begin{equation}
\label{eq:final_loss}
\mathcal{L}=\mathcal{L}_{\mathrm{unmasked}}^{a} + \lambda_1\mathcal{L}_{\mathrm{unmasked}}^{g} + \lambda_2\mathcal{L}_{\mathrm{masked}}^{ma} + \lambda_3\mathcal{L}_{\mathrm{masked}}^{g}.
\end{equation}
Note that we do not apply a multi-view version for the geometry diffusion prior. This is because, experimentally, we found that it contributes less while requiring more time cost.

\section{Experiments}
\subsection{Experimental Setup}

\noindent \textbf{Datasets.}
We collect two real-world datasets for the experiments, called \textit{Real-S} and \textit{Real-L}. Next, we provide a detailed description of both datasets.

\noindent \textit{Real-S}. This dataset comprises all $10$ real-world scenes with slight viewpoint variations from~\cite{mirzaei2023spin}. In each scene, there are $60$ training RGB images that include the unwanted object and the corresponding human-annotated masks. Additionally, $40$ test images without the object are provided for quantitative evaluations. When assessing depth quality, we adhere to a standard scheme~\cite{wang2023sparsenerf} of training a robust NeRF model exclusively on test views to produce pseudo-ground-truth depth maps.
It is also important to note that both RGB and depth images have been resized to $1008 \times 567$, aligning with the dimensions used in~\cite{mirzaei2023spin}.

\noindent \textit{Real-L}. This dataset, sourced from~\cite{weder2023removing}, comprises $16$ scenes with large viewpoint variations. They offer a wide range of diversity, encompassing differences in background texture, object size, and scene geometry.
For each scene, two RGB-D sequences are available. One sequence includes the object, while the other does not, facilitating comprehensive evaluation and analysis. Since the depth maps are provided, we do not need to generate pseudo-ground-truth depth maps.
Also, note that the RGB-D images have been resized to $192 \times 256$.

\noindent \textbf{Metrics.}
To assess inpainting quality, we compare the output novel-view images generated by different approaches to the corresponding ground-truth images for each test image. Specifically, for \textit{Real-S}, all metrics in the paper are exclusively calculated within the bounding boxes of masked regions, while for \textit{Real-L}, due to the limited input resolution, we directly evaluate the full image.
For appearance evaluation, we employ three standard metrics: PSNR, LPIPS, and FID. To assess geometric recovery, we compute the $L_2$ errors between the depth maps rendered by our system and the (pseudo) ground-truth depth maps. Observing that the masking scheme (whether the unmasked region is set to 0 or GT) and the LPIPS version (VGG or Alex) can affect the results, we provided more results in the supplementary file.

\noindent \textbf{Parameters.} 
We implemented our NeRF inpainting model built upon SPIn-NeRF~\cite{mirzaei2023spin} and trained it on $4$ NVIDIA V100 GPUs for $10,000$ iterations using the Adam optimizer with a learning rate of $10^{-4}$. 
As for the diffusion prior, the size of all latent inputs is set to $256\times256$. We set the range of timesteps as $t_{\tt{min}} = 0.02$ and $t_{\tt{max}} = 0.98$. In addition, we implement an annealing timestep scheduling strategy~\cite{zhu2023hifa}, which allocates more training steps to lower values of $t$. For the classifier-free guidance (CFG), we choose values within the range of $[7.5, 25]$ for $\mathcal{L}_{\mathrm{masked}}^{ma}$ and $[2.5, 7.5]$ for $\mathcal{L}_{\mathrm{masked}}^{g}$. The number of sampled views $N$ is set to 5 in all cases. When rendering a single view, we select its four nearby views to calculate the multi-view score.
Lastly, for the balance weights in Eq.~\ref{eq:final_loss}, we empirically set $\lambda_1 = 0.1$ and $\lambda_2 = \lambda_3 = 0.0001$. 

\noindent \textbf{Training with high-resolution images.} 
To enable training on high-resolution images, such as those with dimensions like $1008\times567$, we employ a separate batching scheme. In each iteration, we randomly select $1024$ rays from the unmasked regions across all training views to reconstruct those areas. Given the context-sensitive nature of the text-to-image diffusion model, for the masked region, we select all rays within a single image that correspond to the masked regions. Following this, we combine the rendered colors from these rays with the unmasked pixels to create a complete image, which is subsequently fed into the diffusion model for further processing.

\begin{figure*}[t]
  \centering
  \includegraphics[width=\linewidth]{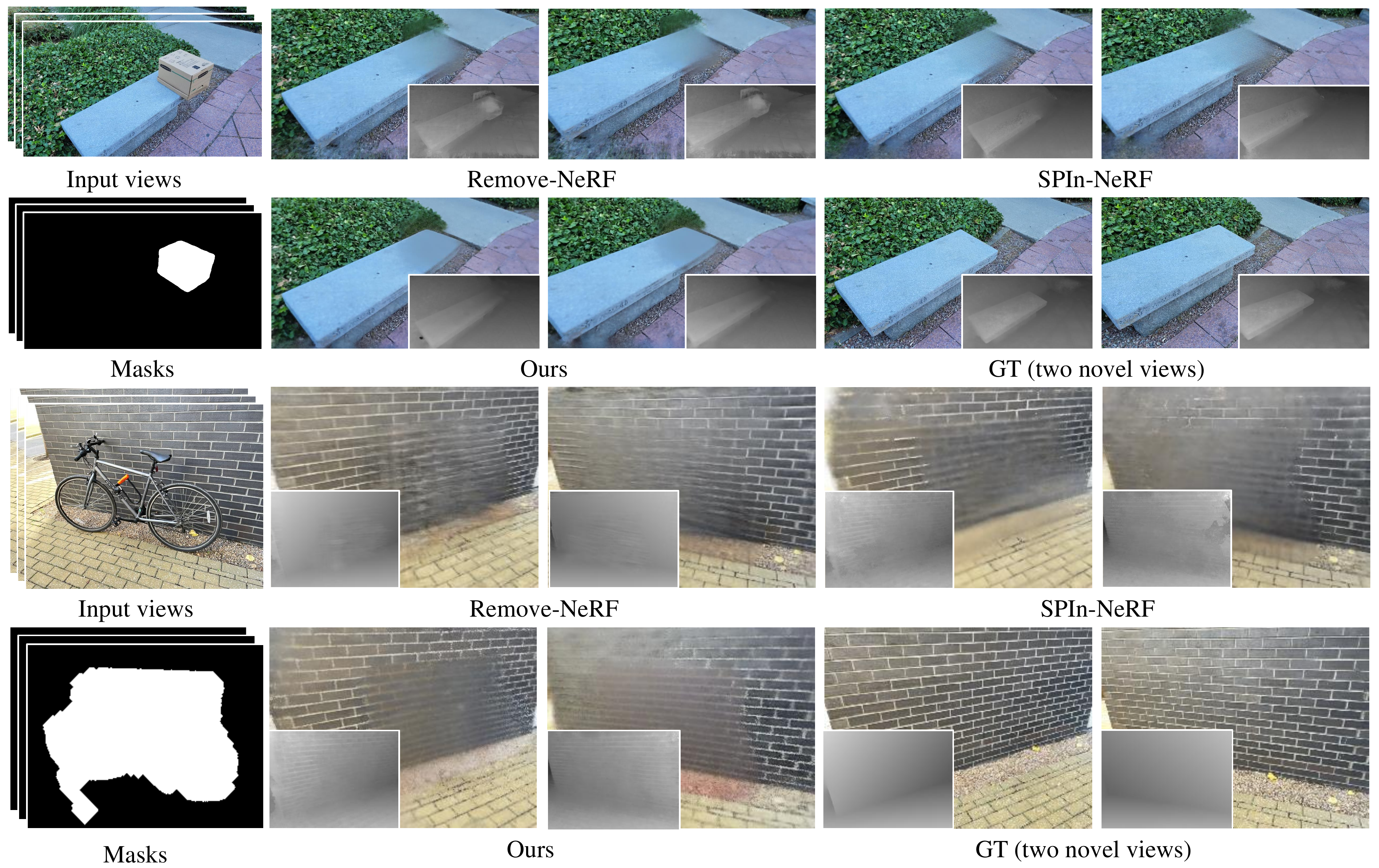}
\caption{Visual comparison with two representative approaches \cite{mirzaei2023spin,weder2023removing} on two scenes. The first scene is from the \textit{Real-S} dataset with accurate masks, while the latter is from the \textit{Real-L} dataset with large, roughly-covered masks. In the first scene, the input text prompt is \textit{``A stone bench''} and for the second scene, it is \textit{``A brick wall''}. Our method effectively handles both types of scenes, successfully generating view-consistent scenes with valid geometries (see the bench shape) and detailed textures (see the brick seam).}
  \label{fig:results_example1}
\end{figure*}

\subsection{Results}
\noindent \textbf{Baselines.} Considering the superior performance demonstrated by recent NeRF inpainting methods \cite{mirzaei2023spin, weder2023removing} compared to traditional video and image-based inpainting pipelines, our focus is primarily on approaches that leverage the foundational NeRF representation. In total, we compare two NeRF inpainting approaches: SPIn-NeRF~\cite{mirzaei2023spin} with LaMa~\cite{suvorov2022resolution}, 
and Remove-NeRF~\cite{weder2023removing} with LaMa~\cite{suvorov2022resolution}. 
Although the two baselines have provided several evaluation results on their datasets, since they both require LaMa inpainting results, we re-executed their released code and reported the results accordingly to ensure a fair comparison. Also, it is noteworthy that we employ LaMa~\cite{suvorov2022resolution} as a 2D inpainter instead of Stable Diffusion. This choice is based on the former's superior quantitative performance~\cite{rombach2022high,mirzaei2023reference}.

\begin{table*}[t]
    \centering
    \caption{\textbf{Comparison with state-of-the-art methods on two real-world datasets.} Our method is \textbf{best} compared to other novel-view synthesis baselines in inpainting the missing regions of the scene. Columns show the deviation from known ground-truth RGB images or depth maps of the scene (without the target object), based on the peak signal-to-noise ratio (PSNR), perceptual metric (LPIPS), feature-based statistical distance (FID), and pixel-wise $L_2$ depth errors.}
    \footnotesize
    \resizebox{1.0\textwidth}{!}{
        \begin{tabular}{l cccc cccc}
            \toprule
            ~ & \multicolumn{4}{c}{\textit{Real-S}} & \multicolumn{4}{c}{\textit{Real-L}} \\
            \cmidrule(lr){2-5}
            \cmidrule(lr){6-9}
            ~ & PSNR$\uparrow$ & LPIPS$\downarrow$ & FID$\downarrow$ & Depth $L_2\downarrow$
              & PSNR$\uparrow$ & LPIPS$\downarrow$ & FID$\downarrow$ & Depth $L_2\downarrow$
            \\
            \midrule 
            Remove-NeRF + LaMa~\cite{weder2023removing}	
            & 17.556 & 0.665 & 254.345 & 8.748 & 25.176 & 0.187 & \textbf{88.245}  & 0.038\\
            SPIn-NeRF + LaMa~\cite{mirzaei2023spin} 
            & 17.466 & 0.574 & \textbf{239.990} &	1.534	&	 25.403 & 0.215 & 103.573	& 0.090 \\
            Ours
            & \textbf{17.667} & \textbf{0.507} & 255.514 & 	\textbf{1.499} & \textbf{25.690} & \textbf{0.181} & 100.452 & \textbf{0.021}	\\
            \bottomrule
        \end{tabular}}
    \label{tab:baseline}
\end{table*}

\begin{table*}[t]
    \centering
    \caption{\textbf{Ablation analysis.} Our method is \textbf{best} compared to different variants of our method in inpainting the missing regions of the scene. Columns show the deviation from known ground-truth RGB images or depth maps of the scene (without the target object). By ablating each component of our approach, we can observe a clear overview of their individual contributions.}
    \footnotesize
    \resizebox{1.0\textwidth}{!}{
        \begin{tabular}{l cccc cccc}
            \toprule
            ~ & \multicolumn{4}{c}{\textit{Real-S}} & \multicolumn{4}{c}{\textit{Real-L}} \\
            \cmidrule(lr){2-5}
            \cmidrule(lr){6-9}
            ~ & PSNR$\uparrow$ & LPIPS$\downarrow$ & FID$\downarrow$ & Depth $L_2\downarrow$
              & PSNR$\uparrow$ & LPIPS$\downarrow$ & FID$\downarrow$ & Depth $L_2\downarrow$
            \\
            \midrule 
(i) ~~(Masked NeRF) 	& 17.121 & 0.761 & 353.766 & 3.089 
& 24.189 & 0.222 & 137.960 &	0.040	\\
(ii) ~(+appearance diffusion)  & 17.020 & 0.587 & 349.066 & 2.574 
&	24.870 & 0.191	&	134.520	&	0.029	\\
(iii) (+inpainted depth map)   & 17.028 & 0.565 & 355.002 &	1.869	
&	24.206 &	0.254	&	138.256  &   0.044	\\
(iv) (+geometry diffusion)   & 17.246 & 0.556 & 346.558  &	1.542
&	25.303 &	0.195	&	129.919  &   0.031	\\
(v) ~(+multi-view) 	& \textbf{17.667} & \textbf{0.507} & \textbf{255.514} & \textbf{1.499} &	\textbf{25.690} & \textbf{0.181} & \textbf{100.452} & \textbf{0.021}		\\
            \bottomrule
        \end{tabular}}
    \label{tab:ablation}
\end{table*}

\noindent \textbf{Quantitative inpainting results}. We conducted a quantitative evaluation to assess the effectiveness of our method compared to the two baselines in terms of both appearance and geometry aspects. The results are reported in Table~\ref{tab:baseline}.

In detail, on the \textit{Real-S} dataset where the view range is limited, all methods yield similar PSNR values. However, when considering the metric of LPIPS, which measures the perceptual quality and realism of the inpainted image, our method excels and demonstrates superiority over existing methods. 
When evaluating the \textit{Real-L} dataset, which features significant view variations, we observed that Remove-NeRF performs better than SPIn-NeRF on the LPIPS metric. This difference in performance can be attributed to the presence of incorrect and inconsistent 2D inpainting results. In such cases, the view selection mechanism of Remove-NeRF can help avoid the incorporation of many incorrect views, resulting in improved performance.
Our MVIP-NeRF achieves superior performance compared to other methods for several reasons. It not only avoids direct dependence on 2D inpainting but also leverages rich knowledge distilled from diffusion prior and shares multiple sampled views. This approach, in turn, enhances performance and contributes to its exceptional results.

\noindent \textbf{Visual inpainting results.} 
Apart from the quantitative comparisons, we also present visual comparisons. 
For each scene, we select two views with relatively large viewing angle deviations to showcase the respective inpainting results. Figure~\ref{fig:results_example1} illustrates the inpainting results for a scene from the \textit{Real-S} dataset, characterized by highly accurate masks, in the first two rows. In contrast, the latter two rows depict results for a scene from the \textit{Real-L} dataset, featuring large masks that roughly cover unwanted areas.
Visual results demonstrate that our method effectively handles both types of scenes, successfully reconstructing view-consistent scenes with detailed textures (as evidenced by the visible brick seams in the second scene) and reasonable geometries (see the completed shape of the bench in the first scene). For more visual results, please refer to the supplemental file.

\noindent \textbf{Ablation study.}
We first demonstrate the impact of ablating various components of our approach. We begin with training a Masked-NeRF (i) as a baseline and progressively incorporate the core modules: (i) Masked-NeRF, namely training a NeRF only using the unmasked pixels; (ii) introducing the appearance diffusion prior; (iii) using explicit depth inpainting results; (iv) replace the explicit depth inpainting prior with our geometry diffusion prior; and (v) employing a multi-view diffusion score function.

Table~\ref{tab:ablation} provides a clear overview of the contribution of each module on the two datasets. 
Comparing (i) and (ii), the notable improvement in the LPIPS metric underscores the effectiveness of the appearance diffusion prior. However, it is worth noting that the PSNR metric may occasionally show less significant improvements due to the blurring effect, which can result in higher PSNR values.
In addition, through a comparison of (ii), (iii), and (iv), we observed that when employing explicit depth inpainting results for NeRF reconstruction, the occasional inaccuracies in the inpainted depth lead to suboptimal geometry recovery. However, by leveraging the geometry diffusion prior, the reconstructed scenes exhibit enhanced geometry quality. Finally, by comparing (iv) and (v), the observed improvement further affirms the effectiveness of the multi-view score.

\begin{figure}[t]
    \centering
    \includegraphics[width=\linewidth]{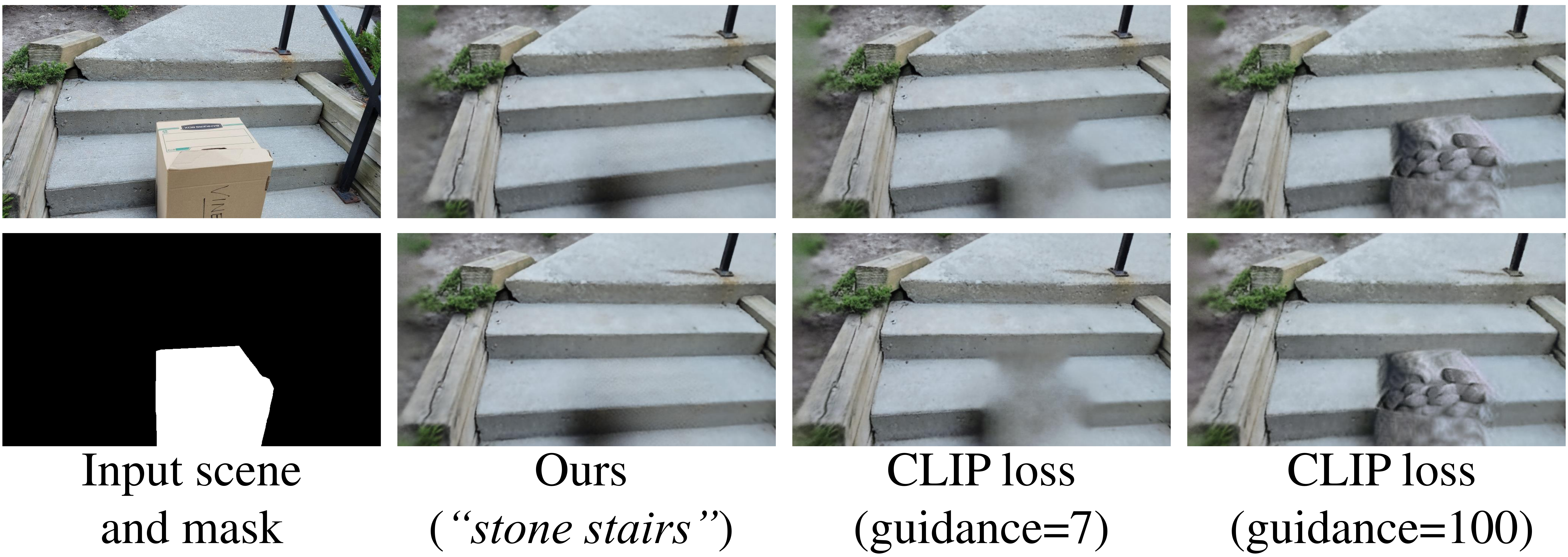}
    \caption{Comparison with CLIP guidances. The input text prompt is \textit{``Stone Stairs''}. For each method, we show two novel view renderings. Our method can faithfully remove the unwanted object and recover the underlying structure.}
    \label{fig:clip}
\end{figure}

\noindent \textbf{Results with CLIP prior.} Interestingly, we found that many previous approaches use the CLIP loss~\cite{radford2021learning} to supervise the alignment between synthesized views and the input text cues. To further exploit the potential prior and validate the effectiveness of our work, we replace the diffusion prior with the CLIP loss, which computes a feature loss between the inpainted image and the given text prompt. As illustrated in Fig.~\ref{fig:clip}, we believe that the CLIP loss is relatively weak, making it challenging to recover the underlying appearance and geometry.

\section{Conclusion}
In this work, we introduce \textit{MVIP-NeRF}, a novel paradigm that harnesses the expressive power of diffusion models for multiview-consistent inpainting on NeRF scenes. Technically, to ensure a valid and coherent recovery of both appearance and geometry, we employ diffusion priors to co-optimize the rendered RGB images and normal maps. To handle scenes with large view variations, we propose a multi-view SDS score function, distilling generative priors from multiple views for consistent visual completion.
We demonstrate the effectiveness of our approach over existing 3D inpainting methods and validate our key ideas by carefully crafting model variants. However, our work has several limitations: (i) the use of diffusion priors for iterative detail recovery affects efficiency, (ii) our method requires effort to tune hyper-parameters of diffusion priors, such as the CFGs, and (iii) as previous work~\cite{mirzaei2023spin,weder2023removing}, our method cannot remove shadows.

\section*{Acknowledgements}
This study is supported by NTU SUG-NAP and under the RIE2020 Industry Alignment Fund - Industry Collaboration Projects (IAF-ICP) Funding Initiative, as well as cash and in-kind contribution from the industry partner(s). It is also supported by Singapore MOE AcRF Tier 2 (MOE-T2EP20221-0011).

\appendix
\clearpage 

\begin{table*}
	\centering
	\caption{\textbf{Comparison with state-of-the-art methods with different 2D inpaiting methods.} Our method is relatively better compared to other novel-view synthesis baselines in inpainting the missing regions of the scene. }
	\footnotesize
	\resizebox{1.0\textwidth}{!}{
		\begin{tabular}{l cccc cccc}
			\toprule
			~ & \multicolumn{4}{c}{\textit{Real-S}} & \multicolumn{4}{c}{\textit{Real-L}} \\
			\cmidrule(lr){2-5}
			\cmidrule(lr){6-9}
			~ & PSNR$\uparrow$ & LPIPS$\downarrow$ & FID$\downarrow$ & Depth $L_2\downarrow$
			& PSNR$\uparrow$ & LPIPS$\downarrow$ & FID$\downarrow$ & Depth $L_2\downarrow$
			\\
			\midrule 
			Remove-NeRF + LaMa~\cite{weder2023removing}	
			& 17.556 & 0.665 & 254.345 & 8.748 & 25.176 & 0.187 & \textbf{88.245}  & 0.038\\
			Remove-NeRF + SD~\cite{weder2023removing}	
			& 17.381 & 0.677 & 245.941 & 9.997 & 24.612 & 0.201 & 110.817 & 0.029\\
			SPIn-NeRF + LaMa~\cite{mirzaei2023spin} 
			& 17.466 & 0.574 & 239.990 &	1.534	&	 25.403 & 0.215 & 103.573	& 0.090 \\
			SPIn-NeRF + SD~\cite{mirzaei2023spin} 
			& 17.497 & 0.604 & \textbf{227.243} &	1.610	&	25.102 & 0.194 & 108.286 & 0.089 \\
			Ours
			& \textbf{17.667} & \textbf{0.507} & 255.514 & 	\textbf{1.499} & \textbf{25.690} & \textbf{0.181} & 100.452 & \textbf{0.021}	\\
			\bottomrule
	\end{tabular}}
	\label{tab:baseline_SD}
\end{table*}

\begin{figure*}[htb]
	\centering
	\includegraphics[width=\linewidth]{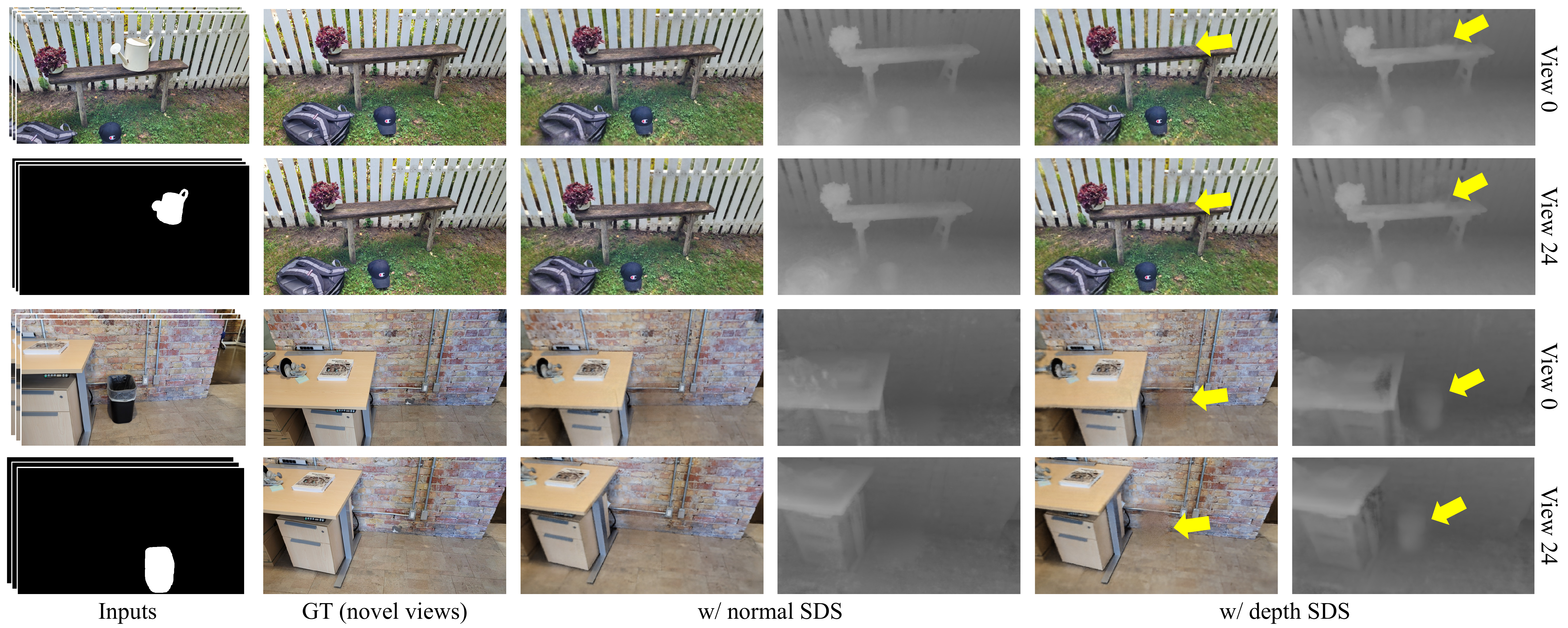}
	\caption{Comparison of depth SDS and our normal SDS. For each scene, we generate RGB images and depth maps for two novel views. Notably, the result of depth SDS reveals limitations in geometry recovery and introduces color distortions. Yellow arrows indicate the less pleasing regions. }
	\label{fig:depth_sds}
\end{figure*}

\begin{figure*}[htb]
	\centering
	\includegraphics[width=\linewidth]{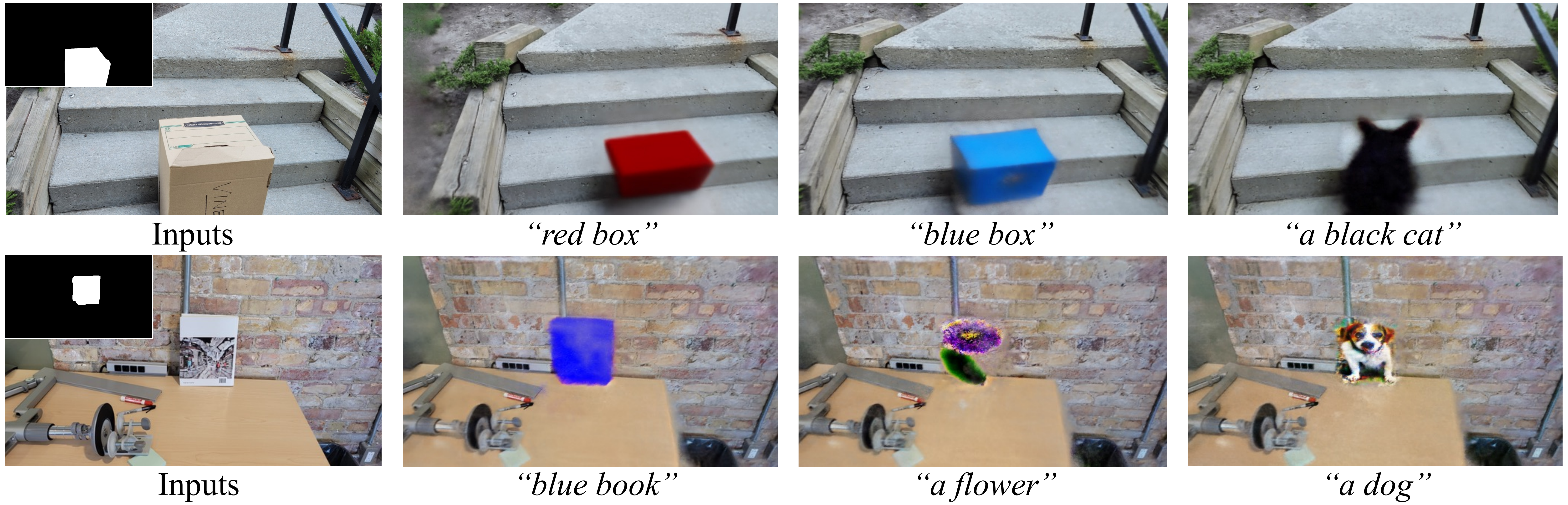}
	\caption{Controlibility of our method. Our method can yield different inpainting results by setting different text prompts.}
	\label{fig:controlibility}
\end{figure*}

In this supplementary material, we provide more detailed information to complement the main manuscript. 
Specifically, we first conduct more ablation studies to analyze our method, by using Stable Diffusion (SD) inpainting results~\cite{rombach2022high} as explicit prior for Remove-NeRF~\cite{weder2023removing} and SPIn-NeRF~\cite{mirzaei2023spin}.
Then, we formulate a depth SDS to further explain why we use the normal map as a geometry representation to distill knowledge from the pre-trained diffusion model.
Next, we provide the additional controllability of our method, more qualitative results, and failure cases. 
Finally, we report the specific parameter configurations utilized in optimizing each NeRF scene from both datasets.  

\section{Comparisons with SD inpainting prior}
Note that both Remove-NeRF and SPIn-NeRF leverage LaMa~\cite{suvorov2022resolution} for independent inpainting across multiple views, followed by the optimization of NeRF scenes. While LaMa has demonstrated superior quantitative performance, as reported in~\cite{rombach2022high,mirzaei2023reference}, we conduct an additional comparison with SPIn-NeRF + SD and Remove-NeRF + SD. As reported in Table~\ref{tab:baseline_SD}, we observe that: i) the SD-based inpainting method may not improve the LaMa-based version, and ii) our method still shows better performance than the SD-based inpainting approaches.

\section{Depth SDS}
As stated in our main paper, we incorporate a geometry diffusion prior to ensure a valid and coherent geometry in the inpainted region. We use normal SDS to distill geometry information from diffusion prior. So, a natural question is: \textit{why not define a depth SDS}? Actually, in our early test, we also formulated the depth SDS as follows:
\begin{align}\label{eq:depth_sds}
	\nabla_\theta \mathcal{L}_{\mathrm{masked}}^{d} = w(t) \big(\boldsymbol{\epsilon}_\phi^{\omega}(\mathbf{z}_t; m,y,t) - \boldsymbol{\epsilon}\big)\frac{
		\partial \mathbf{z}}{\partial \mathbf{d}}\frac{
		\partial \mathbf{d}}{\partial \theta},
\end{align}
where $\mathbf{d}$ denotes the rendered depth map. 

Fig.~\ref{fig:depth_sds} presents the inpainting results with normal SDS and depth SDS. It is evident that the depth SDS result exhibits depth residuals in the inpainted regions, and displays a certain degree of edge distortion in its color images. This observation highlights the challenges associated with depth SDS. We argue that the less satisfactory performance of depth SDS can be attributed to the inherent limitation of depth values in conveying comprehensive geometry information, as opposed to surface normals which more clearly reveal the geometric structures.

\section{Controlibility} 
An additional significant capability of our method is the generation of novel content within the masked region in the 3D scene, which we refer to as controllability. Examples illustrating this capability are presented in Fig.\ref{fig:controlibility}. It is noteworthy that \cite{mirzaei2023reference} also possesses the ability to insert novel content into the 3D scene by providing a different inpainted reference image. In our case, controllability is achieved by supplying a distinct text prompt and a large classifier-free guidance (CFG) value (set to $25$ for all results).

It is essential to acknowledge that, due to our method's reliance on the SDS loss, the generated contents may not exhibit the same level of realism as~\cite{mirzaei2023reference}, which employs realistic inpainted images. The inherent differences in the approaches highlight the trade-offs between controllability and photorealism in content generation within 3D scenes. How to better utilize diffusion to solve this problem would be an interesting direction.

\section{Run time}
As indicated in the main paper, our method, due to its reliance on the diffusion model, requires more time and memory resources. Specifically, for each scene in \textit{Real-S} (image resolution: $1008\times567$), our model can be trained with $2$ v100 GPUs and consumes approximately $6$ hours of computation with $10,000$ iterations. In contrast, Remove-NeRF and SPIn-NeRF, employing a random batching scheme, can operate on a single GPU and complete the task in less than $1$ hour. We can alleviate this problem by feeding the downscaled rendered image into a diffusion model.

\begin{figure*}
	\centering
	\includegraphics[width=\linewidth]{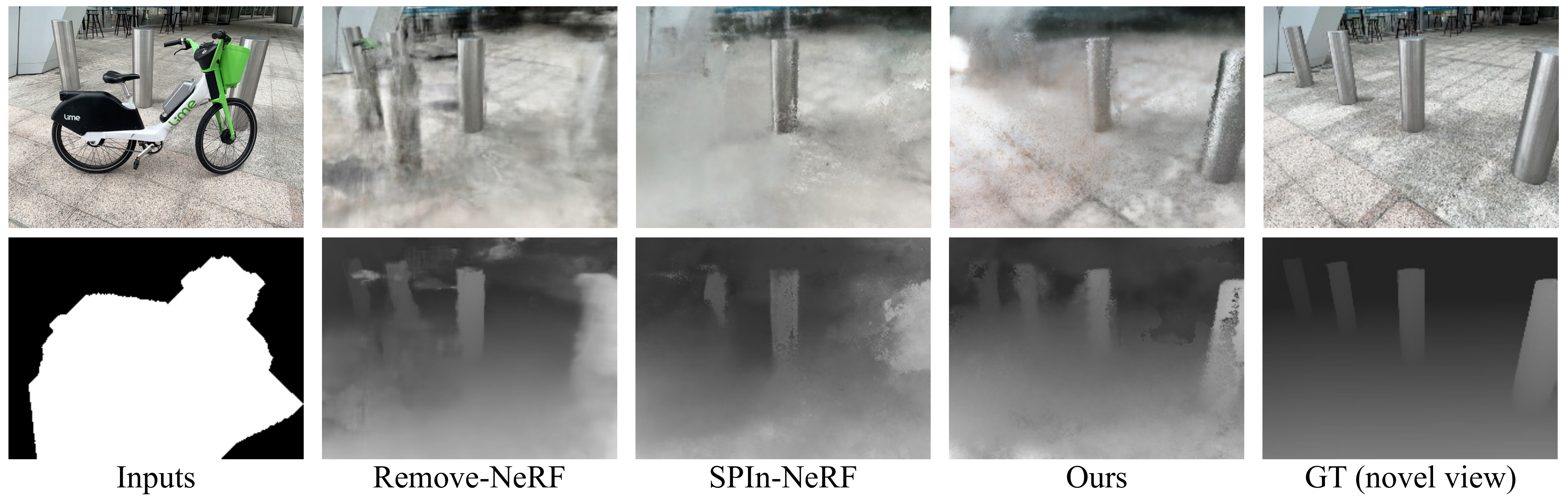}
	\caption{Failure case. The input is a scene with very large undesired areas, and these areas are difficult to adequately describe with a textual prompt (\textit{``a group of metal poles sitting on an outdoor floor''}), our method may exhibit a tendency to generate blurred results. }
	\label{fig:failure_case}
\end{figure*} 

\section{Failure cases}
In scenarios where a scene has very large undesired areas, and these areas are difficult to adequately describe with textual cues, our method may exhibit a tendency to generate blurred results. This limitation arises from the inherent difficulty in capturing fine details or specific features when the inpainting task involves extensive and complex regions that lack clear descriptive cues from external prompts. One potential direction for improvement is to use more accurate masks so that more information can be exploited.

\begin{figure*}
	\centering
	\includegraphics[width=\linewidth]{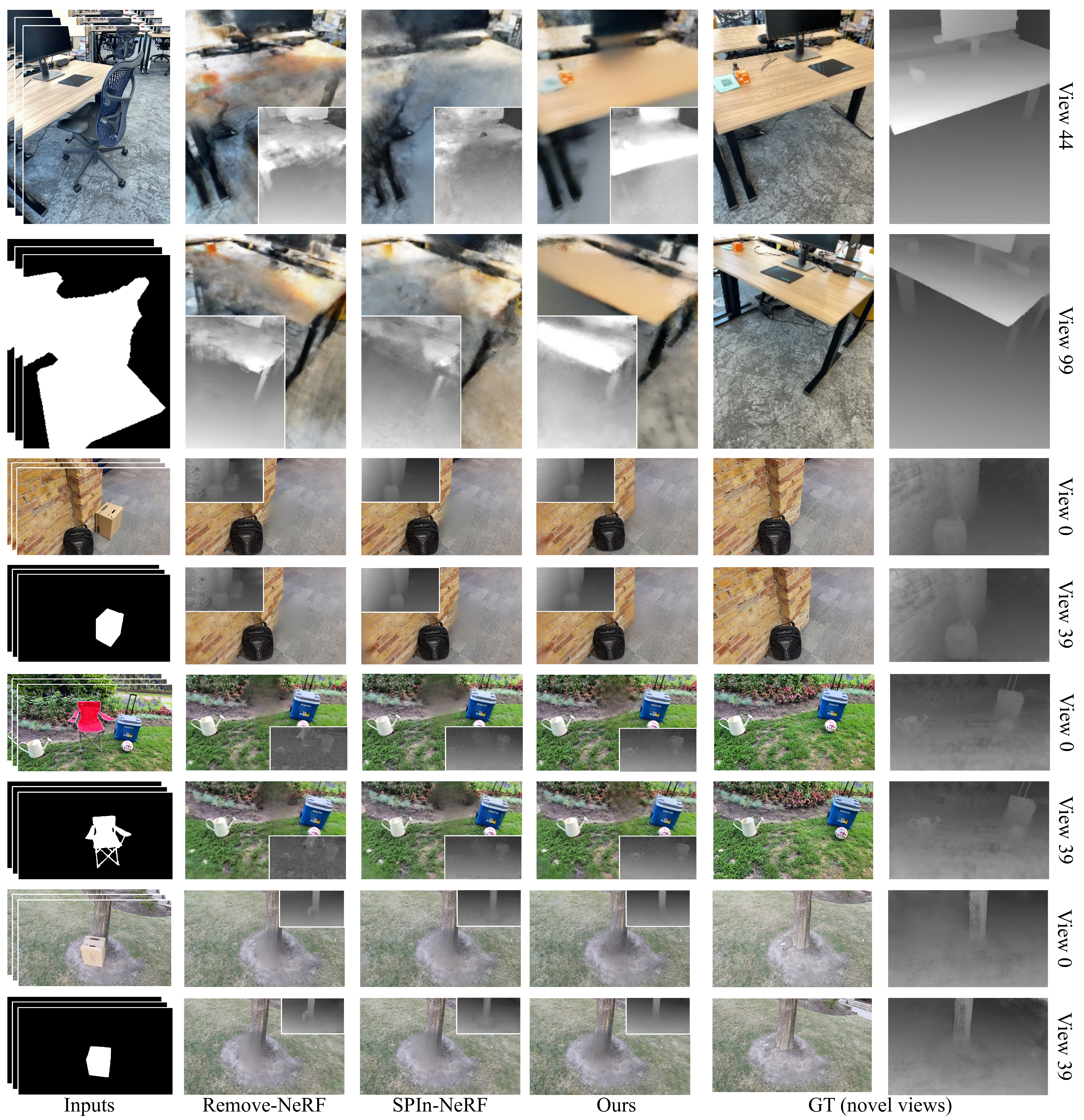}
	\caption{More visual results on different scenes. For each scene, we present inpainted results from two novel viewpoints. It is noteworthy that our approach not only excels in recovering large missing regions but also demonstrates proficiency in restoring intricate textures and maintaining well-aligned geometries.}
	\label{fig:more_results}
\end{figure*} 

\section{More visual results}
More detailed qualitative results for several challenging cases are demonstrated in Fig.~\ref{fig:more_results}. We show the inpainted results from two novel different viewpoints. We can observe that our approach not only excels in recovering large missing regions but also demonstrates proficiency in restoring intricate textures and maintaining well-aligned geometries.

\begin{table*}[h]
	\centering
	\caption{LPIPS results computed by different evaluation schemes. Left side of ``/'': REAL-S. Right side: REAL-L.}
	\setlength{\tabcolsep}{1.5mm}{
		\begin{tabular}{lllll}
			\hline
			& VGG \& umasked=0 & VGG \& umasked=GT & Alex \& umasked=0 & Alex \& umasked=GT \\ \hline
			Remove-NeRF & 0.503/0.123                 & 0.588/0.170                & 0.584/0.121                  & 0.665/0.187                  \\ 
			SPIn-NeRF   & 0.425/0.138                & 0.513/0.212                  & 0.497/0.143                  & 0.574/0.215                    \\ 
			Ours        & 0.409/0.115                 & 0.488/0.163                  & 0.443/0.119                  & 0.507/0.181                     \\ \hline
	\end{tabular}}
	\label{tab:reb:lpips}
\end{table*}

\begin{table*}
	\centering
	\caption{Detailed parameter setting. We report detailed parameters for each scene, including the inputted text prompt, CFG for multi-view SDS ($\mathcal{L}_{\mathrm{masked}}^{ma}$), and CFG for normal SDS ($\mathcal{L}_{\mathrm{masked}}^{g}$). This provides a comprehensive overview of the input configurations}
	\footnotesize
	\resizebox{1.0\textwidth}{!}{
		\begin{tabular}{cllll}
			\toprule
			\multicolumn{1}{l}{}     Dataset &  Scene     & Text prompt                                & CFG-ma & CFG-g \\ \hline
			\multirow{10}{*}{\textit{Real-S}} & 1     & a stone park bench                            & 7.5    & 7.5   \\
			& 2     & a wooden tree trunk on dirt                   & 7.5    & 2.5   \\
			& 3     & a red fence                                   & 7.5    & 7.5   \\
			& 4     & stone stairs                                  & 7.5    & 2.5   \\
			& 7     & a grass ground                                & 15     & 7.5   \\
			& 9     & a corner of a brick wall and a carpeted floor & 12.5   & 5     \\
			& 10    & a wooden bench in front of a white fence      & 7.5    & 7.5   \\
			& 12    & grass ground                                  & 15     & 7.5   \\
			& book  & a brick wall with an iron pipe                & 12.5   & 5     \\
			& Trash & a brick wall                                  & 12.5   & 5     \\
			\hline  
			\multirow{16}{*}{\textit{Real-L}} & 001   & a gray carpet floor                           & 7.5    & 7.5   \\
			& 002   & office desk, carpet floor                     & 25     & 7.5   \\
			& 003   & a sofa                                        & 7.5    & 7.5   \\
			& 004   & black door, white wall, and carpet floor      & 7.5    & 7.5  \\
			& 005   & an office desk                                & 7.5    & 7.5   \\
			& 006   & a stone floor                                 & 7.5    & 7.5     \\
			& 007   & a stone bench                                 & 7.5    & 7.5   \\
			& 008   & a stone wall                                  & 7.5    & 7.5   \\
			& 009   & a wall corner                                 & 7.5    & 7.5     \\
			& 010   & a wall corner and a wooden floor              & 12.5   & 7.5     \\
			& 011   & a white door and a wooden floor               & 12.5   & 7.5     \\
			& 012   & stone staircases                              & 7.5    & 7.5   \\
			& 013   & a wall corner                                 & 12.5   & 7.5   \\
			& 014   & a brick wall corner                           & 12.5   & 7.5     \\
			& 015   & a brick wall                                  & 25     & 7.5     \\
			& 016   & a group of metal poles sitting on an outdoor floor     & 25   & 7.5     \\
			\bottomrule
	\end{tabular}}
	\label{tab:paras}
\end{table*}

\section{Evaluation settings and more quantitative results}
Due to the potential influence of the underlying NeRF architecture, in our evaluation settings, we replaced the unmasked region of the rendered images with their ground truth. Thus, only the masked region contributes to the final error. Observing that the masking scheme (whether the unmasked region is set to 0 or GT) and the LPIPS version (VGG or Alex) can affect the results, we report more detailed results in Table~\ref{tab:reb:lpips}.

\section{Detailed parameter settings}
Given that our method is tailored to leverage the text-to-image diffusion model, we provide detailed parameters for each scene in Table~\ref{tab:paras}, including the inputted text prompt, classifier-free guidance (CFG) for multi-view SDS, and CFG for normal SDS. This provides a comprehensive overview of the input configurations.

{
    \small
    \bibliographystyle{ieeenat_fullname}
    \bibliography{main}

\begin{thebibliography}{39}
\providecommand{\natexlab}[1]{#1}
\providecommand{\url}[1]{\texttt{#1}}
\expandafter\ifx\csname urlstyle\endcsname\relax
  \providecommand{\doi}[1]{doi: #1}\else
  \providecommand{\doi}{doi: \begingroup \urlstyle{rm}\Url}\fi

\bibitem[Chen et~al.(2023)Chen, Chen, Jiao, and Jia]{chen2023fantasia3d}
Rui Chen, Yongwei Chen, Ningxin Jiao, and Kui Jia.
\newblock Fantasia3d: Disentangling geometry and appearance for high-quality
  text-to-3d content creation.
\newblock \emph{arXiv preprint arXiv:2303.13873}, 2023.

\bibitem[Deng et~al.(2022)Deng, Liu, Zhu, and Ramanan]{deng2022depth}
Kangle Deng, Andrew Liu, Jun-Yan Zhu, and Deva Ramanan.
\newblock Depth-supervised nerf: Fewer views and faster training for free.
\newblock In \emph{Proceedings of the IEEE/CVF Conference on Computer Vision
  and Pattern Recognition}, pages 12882--12891, 2022.

\bibitem[Dhariwal and Nichol(2021)]{DiffBeatGAN}
Prafulla Dhariwal and Alexander Nichol.
\newblock Diffusion models beat gans on image synthesis.
\newblock In \emph{NeurIPS}, pages 8780--8794, 2021.

\bibitem[Haque et~al.(2023)Haque, Tancik, Efros, Holynski, and
  Kanazawa]{haque2023instruct}
Ayaan Haque, Matthew Tancik, Alexei~A Efros, Aleksander Holynski, and Angjoo
  Kanazawa.
\newblock Instruct-nerf2nerf: Editing 3d scenes with instructions.
\newblock \emph{arXiv preprint arXiv:2303.12789}, 2023.

\bibitem[Ho et~al.(2020)Ho, Jain, and Abbeel]{DDPM}
Jonathan Ho, Ajay Jain, and Pieter Abbeel.
\newblock Denoising diffusion probabilistic models.
\newblock In \emph{NeurIPS}, 2020.

\bibitem[Kawar et~al.(2022)Kawar, Elad, Ermon, and Song]{kawar2022denoising}
Bahjat Kawar, Michael Elad, Stefano Ermon, and Jiaming Song.
\newblock Denoising diffusion restoration models.
\newblock \emph{Advances in Neural Information Processing Systems},
  35:\penalty0 23593--23606, 2022.

\bibitem[Li et~al.(2023)Li, Ren, Jin, Lan, Wang, Zeng, Wang, and
  Chen]{li2023diffusion}
Xin Li, Yulin Ren, Xin Jin, Cuiling Lan, Xingrui Wang, Wenjun Zeng, Xinchao
  Wang, and Zhibo Chen.
\newblock Diffusion models for image restoration and enhancement--a
  comprehensive survey.
\newblock \emph{arXiv preprint arXiv:2308.09388}, 2023.

\bibitem[Lin et~al.(2023{\natexlab{a}})Lin, Gao, Tang, Takikawa, Zeng, Huang,
  Kreis, Fidler, Liu, and Lin]{lin2023magic3d}
Chen-Hsuan Lin, Jun Gao, Luming Tang, Towaki Takikawa, Xiaohui Zeng, Xun Huang,
  Karsten Kreis, Sanja Fidler, Ming-Yu Liu, and Tsung-Yi Lin.
\newblock Magic3d: High-resolution text-to-3d content creation.
\newblock In \emph{Proceedings of the IEEE/CVF Conference on Computer Vision
  and Pattern Recognition}, pages 300--309, 2023{\natexlab{a}}.

\bibitem[Lin et~al.(2023{\natexlab{b}})Lin, He, Chen, Lyu, Fei, Dai, Ouyang,
  Qiao, and Dong]{lin2023diffbir}
Xinqi Lin, Jingwen He, Ziyan Chen, Zhaoyang Lyu, Ben Fei, Bo Dai, Wanli Ouyang,
  Yu Qiao, and Chao Dong.
\newblock Diffbir: Towards blind image restoration with generative diffusion
  prior.
\newblock \emph{arXiv preprint arXiv:2308.15070}, 2023{\natexlab{b}}.

\bibitem[Liu et~al.(2022)Liu, Shen, Chen, et~al.]{liu2022nerf}
Hao-Kang Liu, I Shen, Bing-Yu Chen, et~al.
\newblock Nerf-in: Free-form nerf inpainting with rgb-d priors.
\newblock \emph{arXiv preprint arXiv:2206.04901}, 2022.

\bibitem[Liu et~al.(2021)Liu, Zhang, Zhang, Zhang, Zhu, and
  Russell]{liu2021editing}
Steven Liu, Xiuming Zhang, Zhoutong Zhang, Richard Zhang, Jun-Yan Zhu, and
  Bryan Russell.
\newblock Editing conditional radiance fields.
\newblock In \emph{Proceedings of the IEEE/CVF international conference on
  computer vision}, pages 5773--5783, 2021.

\bibitem[Mendiratta et~al.(2023)Mendiratta, Pan, Elgharib, Teotia, R~MB,
  GOLYANIK, KORTYLEWSKI, and THEOBALT]{mendiratta2023avatarstudio}
Mohit Mendiratta, Xingang Pan, Mohamed Elgharib, Kartik Teotia, TEWARI~A R~MB,
  V GOLYANIK, A KORTYLEWSKI, and C THEOBALT.
\newblock Avatarstudio: Text-driven editing of 3d dynamic human head avatars.
\newblock \emph{ACM ToG (SIGGRAPH Asia)}, 1:\penalty0 16--18, 2023.

\bibitem[Metzer et~al.(2023)Metzer, Richardson, Patashnik, Giryes, and
  Cohen-Or]{metzer2023latent}
Gal Metzer, Elad Richardson, Or Patashnik, Raja Giryes, and Daniel Cohen-Or.
\newblock Latent-nerf for shape-guided generation of 3d shapes and textures.
\newblock In \emph{Proceedings of the IEEE/CVF Conference on Computer Vision
  and Pattern Recognition}, pages 12663--12673, 2023.

\bibitem[Mildenhall et~al.(2020)Mildenhall, Srinivasan, Tancik, Barron,
  Ramamoorthi, and Ng]{MildenhallSTBRN20}
Ben Mildenhall, Pratul~P. Srinivasan, Matthew Tancik, Jonathan~T. Barron, Ravi
  Ramamoorthi, and Ren Ng.
\newblock Nerf: Representing scenes as neural radiance fields for view
  synthesis.
\newblock In \emph{European Conference on Computer Vision}, pages 405--421.
  Springer, 2020.

\bibitem[Mirzaei et~al.(2022)Mirzaei, Kant, Kelly, and
  Gilitschenski]{mirzaei2022laterf}
Ashkan Mirzaei, Yash Kant, Jonathan Kelly, and Igor Gilitschenski.
\newblock Laterf: Label and text driven object radiance fields.
\newblock In \emph{European Conference on Computer Vision}, pages 20--36.
  Springer, 2022.

\bibitem[Mirzaei et~al.(2023{\natexlab{a}})Mirzaei, Aumentado-Armstrong,
  Brubaker, Kelly, Levinshtein, Derpanis, and
  Gilitschenski]{mirzaei2023reference}
Ashkan Mirzaei, Tristan Aumentado-Armstrong, Marcus~A Brubaker, Jonathan Kelly,
  Alex Levinshtein, Konstantinos~G Derpanis, and Igor Gilitschenski.
\newblock Reference-guided controllable inpainting of neural radiance fields.
\newblock \emph{arXiv preprint arXiv:2304.09677}, 2023{\natexlab{a}}.

\bibitem[Mirzaei et~al.(2023{\natexlab{b}})Mirzaei, Aumentado-Armstrong,
  Derpanis, Kelly, Brubaker, Gilitschenski, and Levinshtein]{mirzaei2023spin}
Ashkan Mirzaei, Tristan Aumentado-Armstrong, Konstantinos~G Derpanis, Jonathan
  Kelly, Marcus~A Brubaker, Igor Gilitschenski, and Alex Levinshtein.
\newblock Spin-nerf: Multiview segmentation and perceptual inpainting with
  neural radiance fields.
\newblock In \emph{Proceedings of the IEEE/CVF Conference on Computer Vision
  and Pattern Recognition}, pages 20669--20679, 2023{\natexlab{b}}.

\bibitem[Oord et~al.(2018)Oord, Li, Babuschkin, Simonyan, Vinyals, Kavukcuoglu,
  Driessche, Lockhart, Cobo, Stimberg, et~al.]{oord2018parallel}
Aaron Oord, Yazhe Li, Igor Babuschkin, Karen Simonyan, Oriol Vinyals, Koray
  Kavukcuoglu, George Driessche, Edward Lockhart, Luis Cobo, Florian Stimberg,
  et~al.
\newblock Parallel wavenet: Fast high-fidelity speech synthesis.
\newblock In \emph{International conference on machine learning}, pages
  3918--3926. PMLR, 2018.

\bibitem[Poole et~al.(2022)Poole, Jain, Barron, and
  Mildenhall]{poole2022dreamfusion}
Ben Poole, Ajay Jain, Jonathan~T Barron, and Ben Mildenhall.
\newblock Dreamfusion: Text-to-3d using 2d diffusion.
\newblock \emph{arXiv preprint arXiv:2209.14988}, 2022.

\bibitem[Radford et~al.(2021)Radford, Kim, Hallacy, Ramesh, Goh, Agarwal,
  Sastry, Askell, Mishkin, Clark, et~al.]{radford2021learning}
Alec Radford, Jong~Wook Kim, Chris Hallacy, Aditya Ramesh, Gabriel Goh,
  Sandhini Agarwal, Girish Sastry, Amanda Askell, Pamela Mishkin, Jack Clark,
  et~al.
\newblock Learning transferable visual models from natural language
  supervision.
\newblock In \emph{International conference on machine learning}, pages
  8748--8763. PMLR, 2021.

\bibitem[Rombach et~al.(2022)Rombach, Blattmann, Lorenz, Esser, and
  Ommer]{rombach2022high}
Robin Rombach, Andreas Blattmann, Dominik Lorenz, Patrick Esser, and Bj{\"o}rn
  Ommer.
\newblock High-resolution image synthesis with latent diffusion models.
\newblock In \emph{Proceedings of the IEEE/CVF conference on computer vision
  and pattern recognition}, pages 10684--10695, 2022.

\bibitem[Saharia et~al.(2022)Saharia, Chan, Saxena, Li, Whang, Denton,
  Ghasemipour, Gontijo~Lopes, Karagol~Ayan, Salimans,
  et~al.]{saharia2022photorealistic}
Chitwan Saharia, William Chan, Saurabh Saxena, Lala Li, Jay Whang, Emily~L
  Denton, Kamyar Ghasemipour, Raphael Gontijo~Lopes, Burcu Karagol~Ayan, Tim
  Salimans, et~al.
\newblock Photorealistic text-to-image diffusion models with deep language
  understanding.
\newblock \emph{Advances in Neural Information Processing Systems},
  35:\penalty0 36479--36494, 2022.

\bibitem[Schuhmann et~al.(2022)Schuhmann, Beaumont, Vencu, Gordon, Wightman,
  Cherti, Coombes, Katta, Mullis, Wortsman, Schramowski, Kundurthy, Crowson,
  Schmidt, Kaczmarczyk, and Jitsev]{Schuhmann2022LAION5BAO}
Christoph Schuhmann, Romain Beaumont, Richard Vencu, Cade Gordon, Ross
  Wightman, Mehdi Cherti, Theo Coombes, Aarush Katta, Clayton Mullis, Mitchell
  Wortsman, Patrick Schramowski, Srivatsa Kundurthy, Katherine Crowson, Ludwig
  Schmidt, Robert Kaczmarczyk, and Jenia Jitsev.
\newblock Laion-5b: An open large-scale dataset for training next generation
  image-text models.
\newblock \emph{ArXiv}, abs/2210.08402, 2022.

\bibitem[Sella et~al.(2023)Sella, Fiebelman, Hedman, and
  Averbuch-Elor]{sella2023vox}
Etai Sella, Gal Fiebelman, Peter Hedman, and Hadar Averbuch-Elor.
\newblock Vox-e: Text-guided voxel editing of 3d objects.
\newblock In \emph{Proceedings of the IEEE/CVF International Conference on
  Computer Vision}, pages 430--440, 2023.

\bibitem[Sohl-Dickstein et~al.(2015)Sohl-Dickstein, Weiss, Maheswaranathan, and
  Ganguli]{Diff15}
Jascha Sohl-Dickstein, Eric~A. Weiss, Niru Maheswaranathan, and Surya Ganguli.
\newblock Deep unsupervised learning using nonequilibrium thermodynamics.
\newblock In \emph{International Conference on Machine Learning}, page
  2256–2265, 2015.

\bibitem[Song et~al.(2021)Song, Sohl-Dickstein, Kingma, Kumar, Ermon, and
  Poole]{ScoreBased}
Yang Song, Jascha Sohl-Dickstein, Diederik~P Kingma, Abhishek Kumar, Stefano
  Ermon, and Ben Poole.
\newblock Score-based generative modeling through stochastic differential
  equations.
\newblock In \emph{International Conference on Learning Representations}, 2021.

\bibitem[Suvorov et~al.(2022)Suvorov, Logacheva, Mashikhin, Remizova, Ashukha,
  Silvestrov, Kong, Goka, Park, and Lempitsky]{suvorov2022resolution}
Roman Suvorov, Elizaveta Logacheva, Anton Mashikhin, Anastasia Remizova,
  Arsenii Ashukha, Aleksei Silvestrov, Naejin Kong, Harshith Goka, Kiwoong
  Park, and Victor Lempitsky.
\newblock Resolution-robust large mask inpainting with fourier convolutions.
\newblock In \emph{Proceedings of the IEEE/CVF winter conference on
  applications of computer vision}, pages 2149--2159, 2022.

\bibitem[Tang et~al.(2023)Tang, Wang, Zhang, Zhang, Yi, Ma, and
  Chen]{tang2023make}
Junshu Tang, Tengfei Wang, Bo Zhang, Ting Zhang, Ran Yi, Lizhuang Ma, and Dong
  Chen.
\newblock Make-it-3d: High-fidelity 3d creation from a single image with
  diffusion prior.
\newblock \emph{arXiv preprint arXiv:2303.14184}, 2023.

\bibitem[Wang et~al.(2022)Wang, Chai, He, Chen, and Liao]{wang2022clip}
Can Wang, Menglei Chai, Mingming He, Dongdong Chen, and Jing Liao.
\newblock Clip-nerf: Text-and-image driven manipulation of neural radiance
  fields.
\newblock In \emph{Proceedings of the IEEE/CVF Conference on Computer Vision
  and Pattern Recognition}, pages 3835--3844, 2022.

\bibitem[Wang et~al.(2023{\natexlab{a}})Wang, Zhang, Abboud, and
  S{\"u}sstrunk]{wang2023inpaintnerf360}
Dongqing Wang, Tong Zhang, Alaa Abboud, and Sabine S{\"u}sstrunk.
\newblock Inpaintnerf360: Text-guided 3d inpainting on unbounded neural
  radiance fields.
\newblock \emph{arXiv preprint arXiv:2305.15094}, 2023{\natexlab{a}}.

\bibitem[Wang et~al.(2023{\natexlab{b}})Wang, Chen, Loy, and
  Liu]{wang2023sparsenerf}
Guangcong Wang, Zhaoxi Chen, Chen~Change Loy, and Ziwei Liu.
\newblock Sparsenerf: Distilling depth ranking for few-shot novel view
  synthesis.
\newblock \emph{arXiv preprint arXiv:2303.16196}, 2023{\natexlab{b}}.

\bibitem[Wang et~al.(2023{\natexlab{c}})Wang, Yue, Zhou, Chan, and
  Loy]{wang2023exploiting}
Jianyi Wang, Zongsheng Yue, Shangchen Zhou, Kelvin~CK Chan, and Chen~Change
  Loy.
\newblock Exploiting diffusion prior for real-world image super-resolution.
\newblock In \emph{arXiv preprint arXiv:2305.07015}, 2023{\natexlab{c}}.

\bibitem[Wang et~al.(2023{\natexlab{d}})Wang, Yu, and Zhang]{wang2023ddnm}
Yinhuai Wang, Jiwen Yu, and Jian Zhang.
\newblock Zero-shot image restoration using denoising diffusion null-space
  model.
\newblock \emph{The Eleventh International Conference on Learning
  Representations}, 2023{\natexlab{d}}.

\bibitem[Wang et~al.(2023{\natexlab{e}})Wang, Lu, Wang, Bao, Li, Su, and
  Zhu]{wang2023prolificdreamer}
Zhengyi Wang, Cheng Lu, Yikai Wang, Fan Bao, Chongxuan Li, Hang Su, and Jun
  Zhu.
\newblock Prolificdreamer: High-fidelity and diverse text-to-3d generation with
  variational score distillation.
\newblock \emph{arXiv preprint arXiv:2305.16213}, 2023{\natexlab{e}}.

\bibitem[Weder et~al.(2023)Weder, Garcia-Hernando, Monszpart, Pollefeys,
  Brostow, Firman, and Vicente]{weder2023removing}
Silvan Weder, Guillermo Garcia-Hernando, Aron Monszpart, Marc Pollefeys,
  Gabriel~J Brostow, Michael Firman, and Sara Vicente.
\newblock Removing objects from neural radiance fields.
\newblock In \emph{Proceedings of the IEEE/CVF Conference on Computer Vision
  and Pattern Recognition}, pages 16528--16538, 2023.

\bibitem[Yang et~al.(2021)Yang, Zhang, Xu, Li, Zhou, Bao, Zhang, and
  Cui]{yang2021learning}
Bangbang Yang, Yinda Zhang, Yinghao Xu, Yijin Li, Han Zhou, Hujun Bao, Guofeng
  Zhang, and Zhaopeng Cui.
\newblock Learning object-compositional neural radiance field for editable
  scene rendering.
\newblock In \emph{Proceedings of the IEEE/CVF International Conference on
  Computer Vision}, pages 13779--13788, 2021.

\bibitem[Yin et~al.(2023)Yin, Fu, Yang, and Lin]{yin2023or}
Youtan Yin, Zhoujie Fu, Fan Yang, and Guosheng Lin.
\newblock Or-nerf: Object removing from 3d scenes guided by multiview
  segmentation with neural radiance fields.
\newblock \emph{arXiv preprint arXiv:2305.10503}, 2023.

\bibitem[Zhou et~al.(2023)Zhou, He, Yu, Li, and Li]{zhou2023repaint}
Xingchen Zhou, Ying He, F~Richard Yu, Jianqiang Li, and You Li.
\newblock Repaint-nerf: Nerf editting via semantic masks and diffusion models.
\newblock \emph{arXiv preprint arXiv:2306.05668}, 2023.

\bibitem[Zhu and Zhuang(2023)]{zhu2023hifa}
Joseph Zhu and Peiye Zhuang.
\newblock Hifa: High-fidelity text-to-3d with advanced diffusion guidance.
\newblock \emph{arXiv preprint arXiv:2305.18766}, 2023.

\end{thebibliography}
}


\end{document}